# Debiasing Diffusion Model: Enhancing Fairness through Latent Representation Learning in Stable Diffusion Model

Lin-Chun Huang[†, 1], Ching Chieh Tsao[1], Fang-Yi Su[1], Jung-Hsien Chiang[1,2]

[1] National Cheng Kung University, Tainan 701, Taiwan
[2] Corresponding Author
[†] First Author contributed the most to this research.

**Abstract.** Image generative models, particularly diffusion-based models, have surged in popularity due to their remarkable ability to synthesize highly realistic images. However, since these models are data-driven, they inherit biases from the training datasets, frequently leading to disproportionate group representations that exacerbate societal inequities. Traditionally, efforts to debias these models have relied on predefined sensitive attributes, classifiers trained on such attributes, or large language models to steer outputs toward fairness. However, these approaches face notable drawbacks: predefined attributes do not adequately capture complex and continuous variations among groups. To address these issues, we introduce the Debiasing Diffusion Model (DDM), which leverages an indicator to learn latent representations during training, promoting fairness through balanced representations without requiring predefined sensitive attributes. This approach not only demonstrates its effectiveness in scenarios previously addressed by conventional techniques but also enhances fairness without relying on predefined sensitive attributes as conditions. In this paper, we discuss the limitations of prior bias mitigation techniques in diffusion-based models, elaborate on the architecture of the DDM, and validate the effectiveness of our approach through experiments.

**Keywords:** Bias mitigation, Fairness, Generative models, Diffusion models, Latent Representation Learning

## 1    Introduction

Artificial intelligence (AI) is advancing at an unprecedented pace and has become an integral part of numerous fields, including advertising, education, and entertainment. However, this rapid adoption has raised serious ethical concerns, particularly regarding bias in AI systems. Biased AI tends to excessively focus on sensitive attributes like race, gender, and age, which require special consideration to prevent discrimination. This bias can result in unfair treatment of individuals or groups, reinforce social inequalities, and erode public trust in technology. To address these challenges, ensuring the principles of equity, diversity, and inclusion (EDI) in AI development is crucial.

Image generative models, such as Variational Autoencoders (VAEs) (Kingma and Welling 2014), Generative Adversarial Networks (GANs) (Goodfellow et al. 2020), and stable diffusion models (SDMs) (Rombach et al. 2021) are widely used in domains such as advertising, media, and creative industries, where their outputs reach broad audiences. However, since these models have been shown to be highly data-driven (Kenfack et al. 2021; Jain et al. 2022; Maluleke et al. 2022; Perera and Patel 2023), they tend to inherit and even amplify the biases present in their training datasets. For instance, previous studies (Nicoletti and Bass 2023; Ferrara 2024) have highlighted noticeable biases in text-to-image models such as Stable Diffusion (Rombach et al. 2021) and OpenAI's DALL-E (Ramesh et al. 2021). These models exhibited racial and stereotypical biases in their outputs, often reflecting societal inequalities present in their training data. A prominent example is gender bias in occupational image generation: when prompted to generate images of CEOs, these models predominantly produced images depicting men, reflecting the real-world underrepresentation of women in leadership roles. Such biased outputs can perpetuate harmful stereotypes, reinforce existing social inequalities, and further marginalize underrepresented groups in society.

Among the image generative models mentioned above, SDMs have emerged as the main image generative model due to their ability to synthesize highly realistic and high-quality images (Dhariwal and Nichol 2021; Rombach et al. 2021). In existing bias mitigation methods, including those employing classifier-free guidance (Ho and Salimans 2021) and classifier guidance (Dhariwal and Nichol 2021), have been predominantly proposed, with the objective of steering the model towards fairer outcomes by incorporating external control signals during the generation process (LUCCIONI 2023; Pal et al. 2023; Kim et al. 2024; Lin et al. 2024; Parihar et al. 2024; Shen et al. 2024; Lin et al. 2025; Sahili et al. 2025). However, these methods still have a significant limitation; they heavily rely on predefined sensitive attributes or the guidance models specifically trained to recognize such attributes. This dependency is problematic because many sensitive attributes, such as gender, exhibit complex and continuous variations that cannot be precisely categorized using existing methods.

To overcome this limitation, we propose the Debiasing Diffusion Model (DDM), which integrates an indicator designed to learn latent representations that are independent of sensitive attributes during training. Instead of relying on static conditions, DDM dynamically steers the diffusion process toward fairness in a flexible and generalizable way. This design enables DDM to address fairness challenges in scenarios where sensitive attributes are not used as conditioning variables. When a neutral prompt (e.g., "*A human face*") is given, DDM balances the distribution of generated images across demographic categories. Furthermore, DDM is also effective in scenarios where sensitive attributes are explicitly specified as conditioning variables, as it reduces performance disparities across diverse demographic groups.

In summary, the primary contributions of this work are as follows:

1. We present a thorough analysis of the limitations of existing bias mitigation techniques in diffusion models, emphasizing their dependence on predefined sensitive attributes for conditioning.

2. We propose the DDM, a novel framework that integrates a latent representation learning indicator into the SDM to decouple sensitive attributes from learned representations. In scenarios without sensitive attribute conditioning, DDM ensures a balanced distribution of generated images across demographic groups; when sensitive attributes are explicitly provided, it minimizes performance disparities among these groups.
3. We conducted a series of experiments to validate the effectiveness of the DDM, assess its strengths and limitations, and discuss its broader implications for AI fairness and ethics. Specifically, our experiments were conducted using two distinct base models and two different datasets, with performance evaluations performed on both the generated images and the indicator.

This paper is structured as follows. In Section 2, we review related work, providing an overview of fairness in image generative models. Section 3 presents key fairness definitions relevant to our study. In Section 4, we detail our proposed methodology, followed by a description of our experimental setup and results in Section 5. Finally, we discuss our findings and outline potential future research directions in Section 6.

## 2 Related Work

**Bias Mitigation Methods in Diffusion-Based Models.** Efforts to enhance fairness in diffusion-based models can be broadly classified into two categories: training-based methods and inference-based methods. Training-based methods modify the diffusion process during training to reduce bias. Techniques in (Lin et al. 2024; Lin et al. 2025) incorporate classifiers into the training pipeline to guide the model to more fair results, using labeled data to influence latent representations. Similarly, Shen et al. (2024) employs a classifier to predict equitable results, which are then used to steer the diffusion process.

Inference-based methods, on the other hand, emphasize adjustments made after model training to ensure fairness. LUCCIONI (2023) extends classifier-free guidance by embedding fairness constraints into the training phase, reducing the dependence on external supervision to address bias. Parihar et al. (2024) introduces a distribution guidance method that aligns generated outputs with a predefined target distribution during inference, achieving balanced representation among demographic groups. Kim et al. (2024) utilizes specialized prompt engineering at inference time to minimize bias, tailoring input prompts to encourage unbiased outputs. Sahili et al. (2025) explores the integration of large language models (LLMs) to detect and correct bias in generated samples, leveraging the contextual understanding of LLMs to refine the results. Pal et al. (2023) addresses bias by localizing the means of facial attributes in the latent space of the diffusion model using Gaussian Mixture Models (GMM), thereby enhancing fairness in attribute representation across diverse demographic categories.

Despite their advancements, both training-based and inference-based methods typically rely on predefined sensitive attributes as conditioning variables or on external tools, such as classifiers or LLMs, trained on such attributes to enhance fairness. This dependency restricts their applicability, as predefining sensitive attributes may not always be feasible or comprehensive enough to address all dimensions of bias. In contrast, DDM introduces a novel framework that operates independently of predefined sensitive attributes and external guidance mechanisms, overcoming these limitations by embedding bias mitigation directly into the latent representation learning process.

**Fair Representation Learning.** Fair representation learning, first introduced in (Zemel et al. 2013), aims to guide intermediate representations to reduce a model's dependence on sensitive attributes, thereby promoting fairness. These approaches can be broadly categorized into adversarial methods and other innovative frameworks, each offering distinct strategies to disentangle sensitive information from learned representations in models like diffusion-based systems.

Adversarial methods have gained prominence for their ability to obscure sensitive attributes, as demonstrated in several key works (Xie et al. 2017; Madras et al. 2018; Feng et al. 2019). These works employ adversarial frameworks to mask sensitive information within the learned representation, leveraging a competing objective to enhance fairness. However, since training adversarial networks can be challenging (Elazar and Goldberg 2018), researchers have turned to alternative strategies that directly separate sensitive and non-sensitive information. One approach (Chowdhury and Chaturvedi 2022) applies rate-distortion maximization to minimize the influence of sensitive attributes, while another (Oh et al. 2022) uses distributional contrastive learning to disentangle sensitive and non-sensitive components. Similarly, Sarhan et al. (2020) proposes fair disentangled orthogonal representations, utilizing orthogonal priors to ensure independence between these elements. These examples highlight the diversity of strategies to achieve fair representations.

Works such as (Zemel et al. 2013; Feng et al. 2019) employ encoder-decoder architectures with mean squared error (MSE) loss to preserve maximal information in intermediate representations while promoting fairness. Inspired by these encoder-decoder frameworks, DDM adapts this approach by minimizing the MSE loss between the latent representation and its denoised counterpart, enabling the denoising U-Net to reconstruct essential features for downstream tasks while reducing the influence of sensitive attributes.

## 3 Fairness Definition for Image Generative Models

### 3.1 Fairness Discrepancy for Generative Models

In the context of evaluating fairness in generative models, Fairness Discrepancy (FD) has been widely adopted in prior studies (Choi et al. 2020; Teo et al. 2023; Parihar et al. 2024) to quantify disparities across demographic groups. Given a data point $x$, FD is defined as follows:

$$\text{FD} = \left| \text{E}_{p_{ref}}[p(a|x)] - E_{p_\theta}[p(a|x)] \right|, \tag{1}$$

where $a$ represents different groups associated with sensitive attributes, $p_{ref}$ is the uniform distribution among demographic groups, and $p_\theta$ represents the distribution of sensitive attributes in generated data. When sensitive attributes are not used as conditioning variables in a generation task, this metric evaluates fairness by assessing how closely the distribution of generated images aligns with a uniform distribution. A lower value indicates closer alignment with the uniform distribution, suggesting a fairer outcome.

In practice, the expectations can be evaluated by Monte Carlo averaging. We evaluate FD by directly comparing the proportions of different groups between $p_{ref}$ and $p_\theta$. For example, in a generated face dataset, we assess fairness by measuring the difference in the proportion of male images between $p_{ref}$ and $p_\theta$.

### 3.2 Statistical Parity

In fairness-aware learning, Statistical Parity (SP) is a prominent notion of group fairness, as explored in works such as (Xu et al. 2018; LUCCIONI 2023), and is defined as:

$$P(x|\, a_1) = P(x|\, a_2), \forall a_1, a_2 \in A, \forall x \in \mathcal{D}, \qquad (2)$$

where A denotes the set of sensitive attributes and $x$ denotes a data point in the dataset $\mathcal{D}$. In the context of data generation, SP requires that the probability of generating any given data point $x$ remains the same, regardless of the sensitive attribute values (e.g. gender, race) associated with it.

Moreover, to offer a more intuitive characterization of fairness, previous studies have proposed a reformulated metric - known as Statistical Parity Difference (SPD) - in generative models (Xu et al. 2018; Krchova et al. 2023):

$$\text{SPD}(x; a_1, a_2) = |P(\,x \mid a_1\,) - P(\,x \mid a_2\,)|, \quad \forall a_1, a_2 \in A, \forall x \in \mathcal{D}. \qquad (3)$$

In this formulation, $\text{SPD}(x; a_1, a_2)$ quantifies the difference in the likelihood of generating $x$ when conditioned on two different sensitive attributes, $a_1$ and $a_2$, respectively. Under the assumption that the sensitive attribute groups are balanced in the training data, this formulation is equivalent to the original SP measure. Minimizing the absolute value of $\text{SPD}(x; a_1, a_2)$ for all $x$; a value close to zero indicates that the generative model treats data points similarly, regardless of the sensitive attribute. In practice, we assess this measure by computing the differences in the proportions of various groups within the generated dataset.

Unlike SP and SPD employed in classification tasks, which assess whether a classifier's outcomes are independent of sensitive attributes (Kamiran and Calders 2012; Zemel et al. 2013; Barrio et al. 2020; Caton and Haas 2024), Eq. 2 and Eq. 3 are increasingly applied to generative tasks (Xu et al. 2018; Krchova et al. 2023). In this context, SP and SPD are tailored to evaluate the fairness of generated datasets by ensuring balanced representation across demographic groups.

## 4 Methodology

We propose the DDM, a fine-tuning approach to mitigate bias in SDMs. At its core, DDM incorporates a latent representation learning indicator used exclusively during training and removed during inference. During training, the indicator serves as a regularization signal, encouraging the denoising U-Net to learn latent representations independent of sensitive attributes. As a result, the model minimizes confounding factors between generated outputs and these attributes, leading to a more fair and balanced generation. In Section 4.1 and Section 4.2, we discuss two different fairness issues and the problem setup respectively, in Section 4.3, we detail the model architecture, and in Section 4.4 we present the algorithm for our method.

### 4.1 Two Scenarios of Bias Mitigation

In DDM, we address two distinct fairness scenarios by tailoring the training process to different fairness definitions, both accommodated within our framework. In the FD scenario, DDM targets a low FD value, as defined in Eq. 1, where the model generates images using a single prompt without explicit conditioning on sensitive attributes. The objective is to ensure that the generated images remain unbiased across different groups, even when no explicit group information is provided. In the SPD scenario, the model aims to achieve a low SPD value, as defined in Eq. 3. This scenario applies when the prompt explicitly specifies sensitive attributes, ensuring that the generated distributions across different groups are identical.

### 4.2 Problem Setup

Consider a training dataset comprising a target training dataset $\mathcal{D}_t$ and a non-target training dataset $\mathcal{D}_{nt}$. The target set, labeled 1, consists of images related to our generation target, while the non-target set, labeled 0, includes images unrelated to this target. We tailor the training process to two fairness scenarios using distinct prompt strategies. In the FD scenario, we employ a single prompt $p_t^{(0)}$ without explicit sensitive attributes (e.g., "*A human face*") and train the DDM to ensure unbiased representation across groups. In the SPD scenario, prompts explicitly specify sensitive attributes (e.g., "*A number 0*" or "*A number 1*"). Here, we utilize multiple prompts $p_t^{(0)}, p_t^{(1)}, \dots, p_t^{(N)}$, where $N$ represents the number of prompts, each paired with corresponding attribute conditions to achieve identical distributions across different groups. For the non-target set, we assign a single prompt $p_{nt}$ that describes characteristics distinct from the target class, enabling DDM to differentiate it from the target set during training. Formally, the sets are defined as $\mathcal{D}_t = \left\{ \left( x_i, 1, p_t^{(j)} | j = 0, 1, \dots, N \right) \right\}$ and $\mathcal{D}_{nt} = \{(x_i, 0, p_{nt})\}$, where $x_i$ denotes the $i$-th image in the dataset, and the binary label 0 or 1 indicates whether $x_i$ is a non-target or target image, respectively.

Using the training dataset, the DDM aims to train a model capable of generating a fair generated dataset $\mathcal{D}_{generated}$ with prompts $p_t^{(j)}$'s. In other words, $\mathcal{D}_{generated}$ satisfies either a low FD or SPD value, depending on the specific scenario.

## 4.3 Latent Representation Learning Indicator

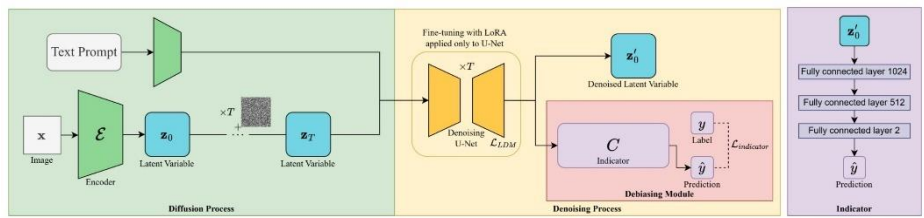

**Fig. 1** DDM architecture. DDM consists of three components: the diffusion process, the denoising process, and the debiasing module. The diffusion and denoising processes follow the original SDM architecture, while the debiasing module, including the indicator, is introduced in our approach. During training, the parameters of the encoder remain fixed, and only those of the denoising U-Net and the indicator are updated. Initially, an image is passed through an encoder to obtain its latent representation. The latent representation is then combined with the encoded prompt, and noise is added. The denoising process subsequently attempts to reconstruct the original latent representation from the noisy latent feature. Finally, the denoised latent feature is passed through the indicator. Note that the design of the indicator, shown in the figure on the right, consists of three fully connected layers and is designed to predict whether a latent representation, which contains image features, belongs to the target or non-target category

**Model Architecture.** Our method is motivated by investigations into fair representation learning methods (Zemel et al. 2013; Feng et al. 2019) and insights from studies on decision-making processes in convolutional neural networks (Baehrens et al. 2010; Zeiler and Fergus 2014; Ribeiro et al. 2016; Lundberg and Lee 2017). Within fair representation learning, previous works (Zemel et al. 2013; Feng et al. 2019) have utilized encoder-decoder architectures to ensure that models preserve image features as comprehensively as possible within the latent representations produced by the encoder while reducing the influence of sensitive attributes in this information to achieve fairness. Concurrently, research on convolutional neural networks (Baehrens et al. 2010; Zeiler and Fergus 2014; Ribeiro et al. 2016; Lundberg and Lee 2017) reveals that these models capture meaningful features from data in a structured and compositional manner rather than relying on random, uninterpretable patterns. Together, these findings inspire our approach to designing a generative framework that enhances fairness and robustness.

Building on these insights, we propose a latent representation learning indicator integrated alongside the denoising U-Net within the DDM, as depicted in Fig. 1 This indicator comprises multiple fully connected layers that process latent representations from the denoising U-Net and is trained to classify them as either target or non-target. Leveraging the architecture of SDM, which adds noise to latent representations and subsequently denoises them, we ensure that the denoised representations retain maximal information. The indicator then guides the model to learn robust features, minimizing the influence of biases associated with sensitive attributes.

During training, both the indicator and the SDM are trained simultaneously. The indicator generates a prediction $\hat{y}$, which is then used to compute the loss:

$$\mathcal{L}_{indicator}(y, \hat{y}) = \sum_{i \in 0,1} y_i \, log\left(\frac{1}{\hat{y}_i}\right). \tag{4}$$

The indicator loss $\mathcal{L}_{indicator}$ is calculated using the cross-entropy between the true labels $y_i$ and the predicted probabilities $\hat{y}_i$, where $i$ denotes whether the latent representation corresponds to the target or non-target class.

Meanwhile, the SDM's reconstruction loss $\mathcal{L}_{SDM}$ can be expressed as:

$$\mathcal{L}_{SDM} \coloneqq E_{\mathbb{E}(x), u, \epsilon \sim \mathbb{N}(0,I), t}\big[|\epsilon - \epsilon_\theta(z_t, t, \tau_\theta(u))|_2^2\big]. \tag{5}$$

$\mathcal{L}_{SDM}$ quantifies the expected squared L2 norm of the difference between the noise $\epsilon$ and the predicted noise $\epsilon_\theta$ given the noised latent variable $z_t$, the time step $t$, and the conditioning encoder $\tau_\theta$. Noise $\epsilon$ is sampled from a standard normal distribution $\mathcal{N}(0, I)$, and $u$ represents conditioning information, such as textual prompts.

This loss function plays a central role in training the SDM, ensuring that the model effectively learns to denoise the latent variable and reconstruct the data. Therefore, the overall loss function of the entire model can be expressed as:

$$\mathcal{L}_{DDM} = (1 - \alpha)\mathcal{L}_{SDM} + \alpha\mathcal{L}_{indicator}, \tag{6}$$

where the hyperparameter α to regulate the balance between $\mathcal{L}_{SDM}$ and $\mathcal{L}_{indicator}$.

**Analysis of Model Effectiveness.** By incorporating the encoder-decoder concept proposed in previous works (Zemel et al. 2013; Feng et al. 2019) and integrating an indicator with the denoising process, we reshape the latent space to mitigate biases, thereby producing fairer generated images.

First, we introduce an encoder-decoder concept that enables the model to learn how to reconstruct the encoder's output through the decoder, thereby allowing the encoder to extract as many features as possible. In this context, the denoising U-Net predicts a clean latent representation $z'_0$ from a noisy latent $z_0$, optimizing the diffusion loss as defined in Eq. 6. This process resembles an encoder-decoder architecture, ensuring that $z'_0$ captures the essential features of $z_0$. This step is critical because it ensures that sensitive attributes are included in the latent representation during training, enabling the indicator to learn how to reduce the model's reliance on these sensitive attributes.

To reduce the model's reliance on sensitive attributes, we further introduce the debiasing module. By assigning identical labels to images with varying sensitive attributes, the indicator is encouraged to prioritize features that are independent of these attributes. Since classifying an image as part of the target dataset requires the model to account for diverse image features, this approach ensures that the learned representations are not dominated by any single sensitive attribute. Ultimately, this acts as a form of regularization, guiding the SDM's training process to diminish reliance on biased features. Therefore, ideally, the classifier should demonstrate consistent performance across diverse demographic groups and exhibit no direct correlation between performance and sensitive attributes, a point we will investigate further in Section 5.3.

## 4.4 Algorithmic Representation for DDM

**Require:** The stable diffusion model $D$, the indicator $C$, the encoder $\mathcal{E}$, the conditioning encoder $\tau_\theta$
1: Initialize weights for the SDM $\theta_D$, the indicator $\theta_C$
2: **repeat**
3:     Sample data pair $(\mathbf{x}, y, u) \sim \mathcal{D}_{training}$,
4:     $\mathbf{z}_0 \leftarrow \mathcal{E}(\mathbf{x})$
5:     Sample time step $t \sim \text{Uniform}(\{1, ..., T\})$
6:     Sample noise $\epsilon \sim \mathcal{N}(0, \mathbf{I})$
7:     $\mathbf{z}'_0 \leftarrow \mathbf{z}_0 + \left[\epsilon - \epsilon_{\theta_D}(\sqrt{\overline{\alpha}_t}\mathbf{z}_0 + \sqrt{1-\overline{\alpha}_t}\epsilon, t, \tau_\theta(u))\right]$
8:     $\hat{y} \leftarrow C(\mathbf{z}'_0)$
9:     Take gradient descent step on $\nabla_{\theta_D}(\|\epsilon - \epsilon_{\theta_D}(\sqrt{\overline{\alpha}_t}\mathbf{z}_0 + \sqrt{1-\overline{\alpha}_t}\epsilon, t, \tau_\theta(u))\|$ and $\nabla_{\theta_{D,C}}(\sum_{i \in 0,1} y \log(\frac{1}{\hat{y}}))$
10: **until** converged

**Fig. 2** Algorithm for DDM training with latent representation learning indicator

In this section, we detail the components and rationale underlying our proposed algorithm for DDM training, as illustrated in Fig. 2. The core innovation lies in incorporating a latent representation learning indicator into the DM framework. This indicator acts as a regularization term that guides the SDM training process, helping to mitigate biases introduced by imbalanced datasets.

Training begins by initializing the parameters for both the SDM $\theta_D$ and the indicator $\theta_C$. These parameters are optimized iteratively to encourage the model to learn latent features that are less correlated with sensitive attributes, thereby mitigating bias and ensuring more balanced feature representation across different groups. This approach mitigates biases by ensuring that the model captures more balanced and representative features across various attributes, regardless of dataset imbalances.

For each training step, a data pair $(x, y, u)$ is sampled from the training dataset, where $x$ represents the input image, $y$ is the corresponding label, and $u$ is the prompt. The image is passed through the encoder $\mathcal{E}$, resulting in the latent variable $z_0$. Following the standard stable diffusion process (Rombach et al. 2021), noise is added to the latent variable, and the model is trained to predict this noise, which is essential for the denoising process in SDMs.

Finally, gradient descent is performed on the total loss function, fine-tuning both $\theta_D$ and $\theta_C$. This effectively guides DDM to focus on features that are less influenced by sensitive attributes, encouraging it to learn more equitable representations.

## 5 Experiments

In this section, we detail the evaluation metrics and a series of experiments conducted to assess the efficacy of our method. In Section 5.1, we introduce the evaluation metrics used to assess the performance of the model in our experiments, in Section 5.2, we detail our experimental design, and in Section 5.3, we present our experimental results.

## 5.1 Evaluation Metrics

In our experiments, we employed five metrics to evaluate the generative performance of our model: Fairness Discrepancy (FD), as defined in Eq. 1 (Choi et al. 2020; Teo et al. 2023; Parihar et al. 2024), Statistical Parity Difference (SPD), as defined in Eq. 3 (Xu et al. 2018; Krchova et al. 2023; LUCCIONI 2023), the Fréchet Inception Distance (FID) (Heusel et al. 2017), the Inception Score (IS) (Salimans et al. 2016), and the unrecognizable proportion. Beyond directly assessing $\mathcal{D}_{generated}$, we also evaluated the indicator's performance across distinct demographic subgroups, further validating the effectiveness of our approach.

**Fréchet Inception Distance.** FID quantifies the similarity between the distributions of real and generated images by comparing statistics. Consequently, it serves as an effective metric for evaluating the image quality of the generated dataset. More specifically, in our evaluations, we calculate FID between the target dataset $\mathcal{D}_t$ and the generated dataset $\mathcal{D}_{generated}$.

**Inception Score.** IS evaluates the quality and diversity of generated images by leveraging the predictive confidence of a pretrained Inception Network (Szegedy et al. 2016). A higher IS value indicates that the generated images belong to well-defined categories while maintaining diversity. In our evaluation, we compute IS using the target dataset $\mathcal{D}_t$ to measure the effectiveness of DDM.

**Unrecognizable Proportion.** To assess the quality of generated images in our experiments, we measure the fraction of samples that fail to be identified as their intended class. Unrecognizable outputs reveal a loss of structural integrity, highlighting the model's inability to generate coherent and recognizable forms. This metric is particularly valuable for assessing generations in which structural clarity is essential, such as distinct shapes or patterns.

**Indicator's Performance across Distinct Demographic Groups.** To assess the indicator's performance across various demographic groups, we prepared a test dataset disjoint from $\mathcal{D}_t$, fed it into the indicator, and computed the mean entropy of each groups. Ideally, since the indicator is designed to reduce the model's reliance on sensitive attributes, the entropy values across distinct demographic groups should exhibit minimal variation and demonstrate no direct correlation between performance and sensitive attributes.

## 5.2 Experimental Setup

**Dataset.** We randomly sample 100 images from each dataset and introduce bias by adjusting the proportions of various groups in $\mathcal{D}_t$. For instance, in the face generation task, one of our experimental setups, we select 80 male images and 20 female images from CelebAMask-HQ (Lee et al. 2020). Similarly, we randomly sample 80 images of one digit and 20 images of another from MNIST (Deng 2012) for the digit generation task. Furthermore, for $\mathcal{D}_{nt}$, we randomly sampled 100 images from the ImageNet (Deng et al. 2009) dataset.

Note that in this study, we adopt binary gender as the sensitive attribute, as the existing gender classification tools classify images exclusively into male and female categories, restricting our analysis to these two classifications. However, this does not imply that we consider gender to be strictly binary. As is widely recognized, gender exists on a spectrum, but no universally agreed-upon categorization method currently exists.

**Prompt Design.** For prompt design, we adopt different strategies for face generation and digit generation experiments. For face generation, we use the prompt "*A human face*" for the target dataset $p_t^{(0)}$, irrespective of sensitive attributes. For the non-target dataset $p_{nt}$, we apply "*Not a human face*" as the prompt. For digit generation, we employ "*A number {number}*" as $p_t^{(i)}$'s, where {number}∈{"0", "1", ..., "9"}. For the non-target dataset, we adopt "*Not a number*" as $p_{nt}$.

**Fine-tunning and Hyperparameter Tuning.** We select two models as our base models and adjust the loss weight α to investigate how different weighting strategies affect the training and final outcomes. For the base models, we use Stable Diffusion 1.5 (SD1.5) and Stable Diffusion 2 (SD2) (Rombach et al. 2021), which we fine-tune using LoRA (Hu et al. 2021) with rank 8. Regarding the loss weight $\alpha$, we experiment with three different settings. In the face generation task, we set $\alpha$ to 0, 0.01, and 0.05, while in the digit generation task, we set $\alpha$ to 0, 0.005, and 0.01, including a baseline configuration where α = 0.

**Evaluation.** In the face generation task, we generate images using "*A human face*" as $p_t^{(0)}$ with a strength of 0.98 to compute relevant metrics, SPD, FID, and IS, and assess the indicator's performance across demographic groups. Specifically, we use the prompt "*A human face*" to generate 1000 images for the image generation task and manually filter out images that do not contain human faces. These images are then used to compute SPD, FID, and IS. In detail, for SPD, we utilize a high-accuracy facial recognition framework, DeepFace (Serengil and Ozpinar 2021), to estimate the gender distribution of the generated images. We then compute the proportions of male and female images to measure the bias. Furthermore, to evaluate the performance of the indicator in demographic groups, we prepare a test dataset containing 50 male and 50 female images, input these into the indicator, and analyze its behavior across these groups.

In addition, for the digit generation task, we randomly select two digits, $d_1$ and $d_2$, to conduct our experiment. We use the prompt "*A number {number}*" as $p_t^{(0)}$ and $p_t^{(1)}$, where {number}∈{"$d_1$", "$d_2$"} and we generate 500 images for each digit. To evaluate FD and unrecognizable proportion, we measure the proportion of generated images recognized as valid digits and compare the recognition rates of $d_1$ and $d_2$. Specifically, we use a classifier trained on the MNIST dataset, achieving a test accuracy of 99%, to identify the generated digits. We consider the model fair if the overall quality of the generated digits is similar for both classes, and we deem it to have high quality if the unrecognizable proportion is low, as fewer unrecognizable digits indicate better structural consistency. Note that we avoid using FID and IS here, as they focus on distributional similarity and diversity rather than the precise structural recognizability required for this task. Similarly, we also evaluate the performance of the indicator in demographic groups by inputting a test dataset containing 50 digit $d_1$ and 50 digit $d_2$ images into the indicator.

## 5.3 Experimental Results

| Model | Sex Ratio | FD↓ | | |
|---|---|---|---|---|
| | | $\alpha = 0$ | $\alpha = 0.01$ | $\alpha = 0.05$ |
| SD1.5 | 0.25 | 0.21 | 0.19 | **0.11** |
| | 1 | 0.08 | 0.06 | **0.03** |
| | 4 | 0.29 | **0.22** | 0.23 |
| SD2 | 0.25 | 0.33 | 0.30 | **0.05** |
| | 1 | 0.24 | **0.05** | 0.09 |
| | 4 | 0.28 | **0.22** | **0.22** |

**Table 1** The FD values under different configurations. This table presents the performance of models trained under different configurations, considering different training datasets and base models while adjusting the loss weight $\alpha$. Specifically, the composition of $\mathcal{D}_{nt}$ remains consistent across all configurations, while the composition of $\mathcal{D}_t$ is specified by the sex ratios in the table and the configuration with $\alpha = 0$ serves as the baseline. The best performance for each metric is highlighted in bold

| Model | Sex Ratio | FID↓ | | |
|---|---|---|---|---|
| | | $\alpha = 0$ | $\alpha = 0.01$ | $\alpha = 0.05$ |
| SD1.5 | 0.25 | **92.26** | 92.62 | 103.58 |
| | 1 | **100.59** | 102.83 | 108.23 |
| | 4 | 137.77 | **112.78** | 120.95 |
| SD2 | 0.25 | **83.80** | 84.25 | 99.40 |
| | 1 | **88.87** | 91.47 | 110.13 |
| | 4 | **104.28** | 105.88 | 109.39 |

**Table 2** The FID values under different configurations. This table presents the performance of models trained under different configurations, considering different training datasets and base models while adjusting the loss weight $\alpha$. Specifically, the composition of $\mathcal{D}_{nt}$ remains consistent across all configurations, while the composition of $\mathcal{D}_t$ is specified by the sex ratios in the table and the configuration with $\alpha = 0$ serves as the baseline. The best performance for each metric is highlighted in bold

| Model | Sex Ratio | IS↑ | | |
|---|---|---|---|---|
| | | $\alpha = 0$ | $\alpha = 0.01$ | $\alpha = 0.05$ |
| SD1.5 | 0.25 | 3.424±0.441 | 3.314±0.238 | **3.651**±0.343 |
| | 1 | 3.270±0.231 | 3.527±0.283 | **3.580**±0.270 |
| | 4 | **4.199**±0.324 | 3.818±0.277 | 3.600±0.333 |
| SD2 | 0.25 | 2.553±0.240 | **2.624**±0.310 | 2.449±0.148 |
| | 1 | 2.590±0.264 | **2.680**±0.228 | 2.567±0.283 |
| | 4 | 3.110±0.273 | **3.138**±0.272 | 2.971±0.202 |

**Table 3** The IS values under different configurations. This table presents the performance of models trained under different configurations, considering different training datasets and base models while adjusting the loss weight $\alpha$. Specifically, the composition of $\mathcal{D}_{nt}$ remains consistent across all configurations, while the composition of $\mathcal{D}_t$ is specified by the sex ratios in the table and the configuration with $\alpha = 0$ serves as the baseline. The best performance for each metric is highlighted in bold

**Face Generation Task.** Our evaluations on $\mathcal{D}_{generation}$ are presented in Table 1, Table 2, and Table 3. In the face generation task, we focus on the FD scenario, aiming to ensure that the generated images exhibit balanced representations across sensitive attribute groups. We selected gender as the primary focus of our evaluation since achieving gender balance poses a highly complex challenge for the model and we believe that demonstrating strong performance in balancing gender is sufficient to validate the effectiveness of our approach.

We adjust the gender ratio within $\mathcal{D}_t$ to introduce artificial bias and assess the effectiveness of our DDM in mitigating it during training. As shown in Table 1, at a sex ratio of 1, both SD1.5 and SD2 exhibit elevated FD values, indicating inherent bias in the pretrained models. After applying our method, FD decreases significantly, demonstrating substantial bias reduction. For datasets with sex ratios 0.25 and 4, while our method does not fully eliminate bias, it consistently reduces FD across all conditions. For FID, as demonstrated in Table 2, DDM generally exhibits a worsening trend across configurations, with SD2 consistently degrading in quality, reflecting a trade-off with enhanced fairness. For IS, as presented in Table 3, SD1.5 improves at balanced and low bias ratios but declines under high bias, while SD2 shows smaller gains and more consistent losses across configurations. This highlights a trade-off between quality and variety: using DDM reduces quality while increasing the range of image variations.

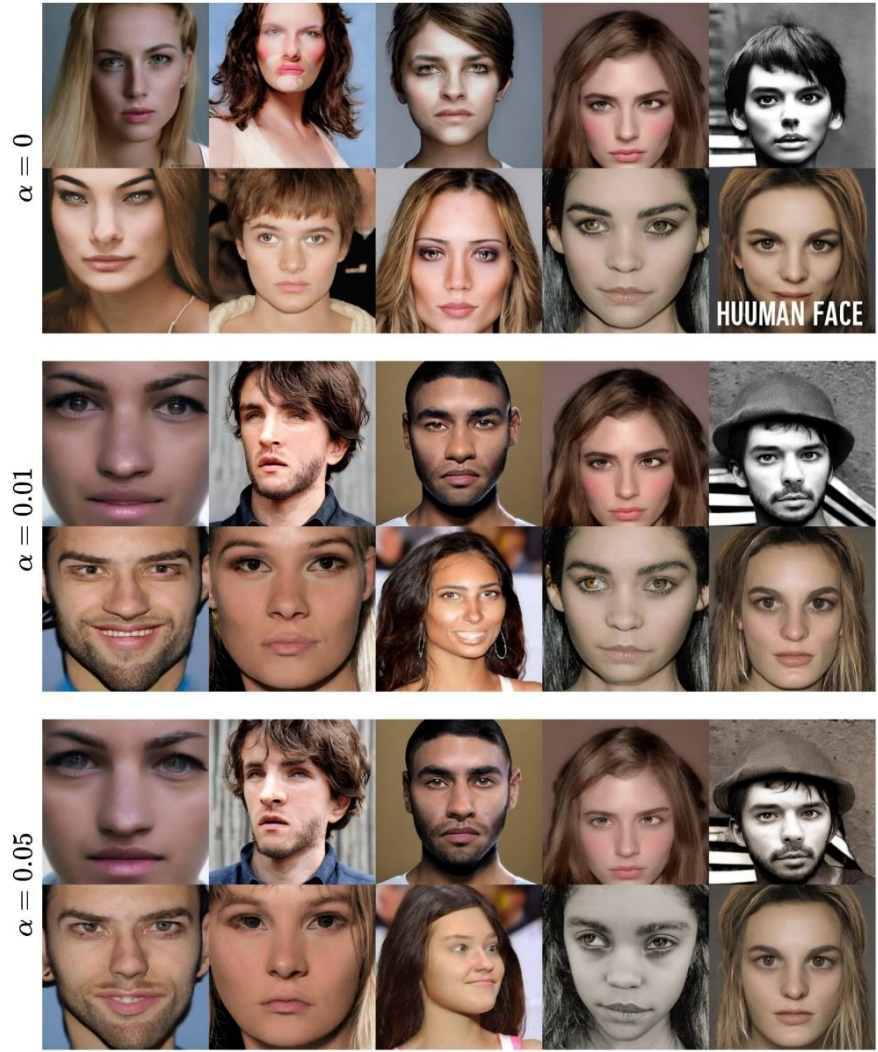

**Fig. 3** The examples of the face generation task. This figure illustrates images generated using the prompt "*A human face*" from a model fine-tuned with SD2 as the baseline. The fine-tuning was performed on a dataset with a sex ratio of 1, setting the parameter α to 0, 0.01, and 0.05. All images were produced using the same set of seeds

Fig. 3 showcases images generated by a model fine-tuned using SD2 as the baseline, trained on a dataset with a sex ratio of 1. The parameter α was adjusted to 0, 0.01, and 0.05. All of these images were produced with the same seed, allowing certain consistent features to be observed across the three different $\alpha$ settings. However, by employing DDM during training, the model's dependence on sensitive attributes is reduced, resulting in different variations in the gender characteristics and thus improving fairness.

Additionally, as observed in this figure, although our method slightly degrades image quality according to the FID metric, this reduction does not lead to differences perceptible to the human eye.

| Model | Sex Ratio | Mean Entropy when $\alpha = 0.01$ | | Mean Entropy when $\alpha = 0.05$ | |
|---|---|---|---|---|---|
| | | Male | Female | Male | Female |
| SD1.5 | 0.25 | 0.6867 | 0.6876 | 0.6903 | 0.6884 |
| | 1 | 0.6876 | 0.6886 | 0.6890 | 0.6891 |
| | 4 | 0.6885 | 0.6886 | 0.6896 | 0.6887 |
| SD2 | 0.25 | 0.6902 | 0.6870 | 0.6866 | 0.6853 |
| | 1 | 0.6896 | 0.6877 | 0.6854 | 0.6893 |
| | 4 | 0.6882 | 0.6882 | 0.6895 | 0.6904 |

**Table 4** The mean entropy of various demographic groups in different configurations. This figure illustrates the performance results of the indicator on a test dataset. Our dataset comprises 50 male images and 50 female images, ensuring no overlap with any training datasets. These images are processed through the DDM, and the output of the denoising U-Net is subsequently fed into the indicator to calculate the average entropy for these images. We adjusted α to 0.01 and 0.05 to collect these statistics, while $\alpha = 0$ was excluded from the analysis since no training of the indicator occurs at this setting

| Model | $d_1$ | $d_2$ | SPD↓ | | |
|---|---|---|---|---|---|
| | | | $\alpha = 0$ | $\alpha = 0.005$ | $\alpha = 0.01$ |
| SD1.5 | 1 | 9 | 0.102 | 0.066 | **0.065** |
| | 3 | 0 | 0.115 | **0.029** | 0.043 |
| | 5 | 4 | 0.053 | 0.031 | **0.004** |
| | 2 | 6 | 0.169 | 0.125 | **0.105** |
| | 8 | 7 | 0.148 | 0.007 | **0.003** |
| SD2 | 1 | 9 | 0.342 | 0.054 | **0.055** |
| | 3 | 0 | 0.118 | 0.153 | **0.047** |
| | 5 | 4 | 0.044 | **0.017** | 0.104 |
| | 2 | 6 | 0.190 | **0.071** | 0.097 |
| | 8 | 7 | 0.081 | 0.054 | **0.007** |

**Table 5** The SPD values under different configurations. This table presents the performance of models trained under different configurations, considering different training datasets and base models while adjusting the loss weight α. Specifically, the composition of $\mathcal{D}_{nt}$ remains consistent across all configurations, while $\mathcal{D}_t$ comprises 80 images of digit $d_1$ and 20 images of digit $d_2$ and the configuration where $\alpha = 0$ serves as the baseline

Table 4 presents the evaluation results for the indicator in the face generation task. Across all model configurations at α = 0.01 and α = 0.05, despite potential biases in the training dataset, the indicator exhibits similar performance on male and female images. To quantify this similarity, we computed the Pearson correlation coefficient between male and female entropy values, yielding $r = -0.063$. This result suggests a negligible linear relationship between gender and entropy, indicating that, during learning, the indicator does not overly rely on sensitive attributes as a basis for classification.

These findings align with our expectations, strengthening the robustness of the indicator in all demographic groups.

| Model | $d_1$ | $d_2$ | Unrecognizable Proportion↓ | | |
|---|---|---|---|---|---|
| | | | $\alpha = 0$ | $\alpha = 0.005$ | $\alpha = 0.01$ |
| | 1 | 9 | 0.64 | 0.54 | **0.52** |
| | 3 | 0 | **0.26** | 0.31 | 0.37 |
| SD1.5 | 5 | 4 | **0.33** | 0.35 | 0.48 |
| | 2 | 6 | **0.47** | 0.66 | 0.68 |
| | 8 | 7 | **0.43** | 0.59 | 0.45 |
| | 1 | 9 | 0.50 | **0.34** | 0.42 |
| | 3 | 0 | 0.42 | 0.38 | **0.32** |
| SD2 | 5 | 4 | **0.27** | 0.39 | 0.34 |
| | 2 | 6 | 0.72 | 0.77 | **0.55** |
| | 8 | 7 | 0.43 | **0.33** | 0.39 |

**Table 6** The unrecognizable proportions under different configurations. This table presents the performance of models trained under different configurations, considering different training datasets and base models while adjusting the loss weight $\alpha$. Specifically, the composition of $\mathcal{D}_{nt}$ remains consistent across all configurations, while $\mathcal{D}_t$ comprises 80 images of digit $d_1$ and 20 images of digit $d_2$ and the configuration where $\alpha = 0$ serves as the baseline

**Digit Generation Task.** Our evaluation results are presented in Table 5 and Table 6. In the digit generation task, we focus on the SPD scenario, aiming to ensure that generated images exhibit balanced representations across specified digit classes. We randomly selected digit pairs $d_1$ and $d_2$ as the primary focus of our evaluation. We adjust the digit ratio within $\mathcal{D}_{target}$ to introduce artificial bias (80 images of $d_1$ and 20 of $d_2$) and assess DDM's ability to mitigate it during training.

For SPD, both SD1.5 and SD2 exhibit substantial improvements with debiasing. In the baseline configuration, both models show elevated SPD values across all pairs of digits, reflecting the inherent bias toward the majority class in the pretrained models. Upon applying our method, SPD consistently decreases across configurations, with SD1.5 achieving larger reductions at higher debiasing strengths, while SD2 shows significant gains with slight adjustments. However, for the proportion of unrecognizable digits, both models show a higher number of unclear samples in several setups. A similar pattern emerges in face generation tasks: as our method improves fairness, it introduces a trade-off between fairness and quality, making it harder for the models to capture the structure of digits or faces, which leads to more unrecognizable outputs.

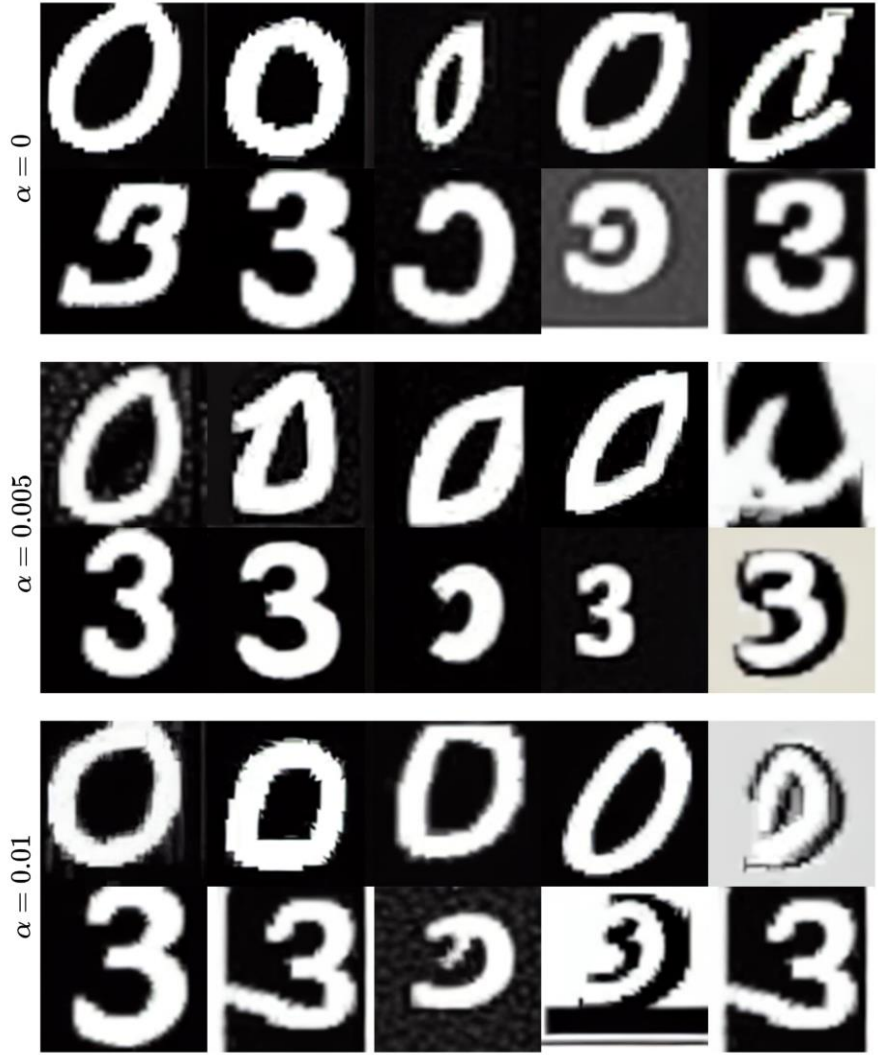

**Fig. 4** The examples of the digit generation task. This figure illustrates images generated using the prompts "*A number 0*" and "*A number 3*" from a model fine-tuned with SD1.5 as the baseline. The fine-tuning was performed on a dataset where $d_1$ represents the digit 3 and $d_2$ represents the digit 0, with the parameter α set to 0, 0.005, and 0.01. All images were produced using the same set of seeds

Fig. 4 presents images generated by a model fine-tuned using SD1.5 as the baseline, trained on a dataset where we select $d_1$ to be digit 3 and $d_2$ to be digit 0. The parameter α was set to 0, 0.005, and 0.01. We randomly sampled 10 images per setting using the same seed, revealing that at α = 0, the model performs significantly worse on digit 3 compared to digit 0. This suggests an overemphasis on learning the digit 0 during training. In contrast, applying DDM slightly degrades the performance on digit 0, introducing some structural imperfections, but improves attention to digit 3. As a result, the generated distribution $\mathcal{D}_{generated}$ achieves greater fairness in performance across both digits.

| Model | $d_1$ | $d_2$ | Mean Entropy when $\alpha = 0.005$ | | Mean Entropy when $\alpha = 0.01$ | |
|---|---|---|---|---|---|---|
| | | | $d_1$ | $d_2$ | $d_1$ | $d_2$ |
| | 1 | 9 | 0.6854 | 0.6802 | 0.6960 | 0.6828 |
| | 3 | 0 | 0.6837 | 0.6828 | 0.6854 | 0.6852 |
| SD1.5 | 5 | 4 | 0.6835 | 0.6826 | 0.6847 | 0.6815 |
| | 2 | 6 | 0.6840 | 0.6796 | 0.6848 | 0.6807 |
| | 8 | 7 | 0.6833 | 0.6754 | 0.6858 | 0.6796 |
| | 1 | 9 | 0.6892 | 0.6873 | 0.6871 | 0.6824 |
| | 3 | 0 | 0.6875 | 0.6862 | 0.6862 | 0.6867 |
| SD2 | 5 | 4 | 0.6870 | 0.6855 | 0.6864 | 0.6798 |
| | 2 | 6 | 0.6869 | 0.6845 | 0.6890 | 0.6865 |
| | 8 | 7 | 0.6845 | 0.6795 | 0.6876 | 0.6828 |

**Table 7** The mean entropy of various demographic groups in different configurations. This figure illustrates the performance results of the indicator on a test dataset. Our dataset comprises 50 images per digit, ensuring no overlap with the training dataset. These images are processed through the DDM, and the output of the denoising U-Net is subsequently input into the indicator to calculate the average entropy. We adjusted α to 0.005 and 0.01 to collect these statistics, while α = 0 was excluded from the analysis since no training of the indicator occurs at this setting

Table 7 presents the evaluation results for the indicator in the digit generation task. Across all model configurations at α = 0.005 and α = 0.01, the indicator demonstrates highly comparable performance on $d_1$ and $d_2$ images. To assess this consistency, we calculated the Pearson correlation coefficient between the entropy values of $d_1$ and $d_2$, obtaining $r = 0.432$. This modest correlation, alongside minimal performance variations, indicates that the indicator's entropy remains substantially stable across digit categories, underscoring its reliability in the digit generation task.

## 6 Discussion

In this section, we discuss the advantages and limitations of DDM by analyzing the contributions our method can offer and the considerations that need attention when applying it.

**Bias Mitigation in Two Scenarios.** DDM effectively reduces bias in pretrained generative models, improving fairness across the FD scenario and the SPD scenario. In the FD scenario, DDM ensures balanced outcomes when sensitive attributes are not specified, dynamically adjusting the diffusion process to promote equitable representations across diverse groups. In the SPD scenario, DDM enhances fairness when sensitive attributes are explicitly conditioned, minimizing performance disparities across demographic categories. Our experiments substantiate this by directly validating $\mathcal{D}_{generated}$ and providing indirect evidence through the indicator's performance across different demographic groups. Although DDM cannot completely eliminate bias due to the inherent complexities of data-driven models, it provides a novel perspective, offering a flexible and innovative approach to advancing equity in generative technologies.

**The Fairness-Quality Trade-off.** Despite these fairness gains, a notable limitation of DDM is the trade-off between fairness and quality, as increasing the debiasing strength $\alpha$ often compromises the quality of generated images. First, a larger $\alpha$ forces the model to prioritize optimizing the fairness indicator during training, which shifts focus away from effectively learning the SDM. Intuitively, this imbalance results in images with indistinct or unrealistic features, as the model neglects structural precision to enforce parity across classes. This degradation in image quality can introduce unpredictable biases in downstream evaluations, undermining the reliability of generated outputs. Second, our method inherently reduces the model's learning of sensitive attributes to mitigate bias, which can impair its ability to restore certain details during inference accurately. Consequently, some image components may not be well-reconstructed, further exacerbating quality loss. These challenges highlight a fundamental tension: while our DDM enhances fairness, it risks sacrificing generation fidelity, a trade-off that requires careful consideration in practical applications.

**Impact of $\alpha$.** Tuning the debiasing parameter $\alpha$ presents a significant challenge for DDM. As shown in Section 5.3, increasing $\alpha$ generally reduces bias, yet deriving a precise formula for an optimal $\alpha$ remains elusive. Excessively high $\alpha$ values can degrade fairness, potentially reversing the intended effect. This issue is compounded by our fairness indicator's reliance on denoised images as input: incomplete denoising leaves residual noise in the latent variables, and an overly large $\alpha$ in such cases may force a well-trained model to produce noisy, low-quality images. This sensitivity underscores the need for careful $\alpha$ calibration in practice and highlights opportunities for future research into adaptive techniques that dynamically adjust debiasing strength to optimize performance across varying conditions.

**Challenges in dataset preparation.** The dataset requirements for our method present a notable drawback. Since the training dataset is composed of both $\mathcal{D}_t$ and $\mathcal{D}_{nt}$, this necessitates the preparation of a dataset effectively double in size, which in turn demands more extensive training efforts. However, this challenge can be mitigated by employing fine-tuning techniques such as LoRA (Hu et al. 2021), which can expedite the training process. Additionally, fine-tuning typically requires a smaller dataset compared to training from scratch. In our experiments, we successfully fine-tuned a high-quality model with a total of only 200 images, demonstrating the feasibility of refining a pre-existing model without an extensive dataset.

**Conclusion.** In this study, we introduced the DDM, a novel framework that enhances fairness in SDMS through latent representation learning, independent of predefined sensitive attributes. Our results show that DDM effectively reduces bias in both FD and SPD scenarios, providing a flexible approach to equitable generative technologies. Despite these advances, the fairness-quality trade-off underscores the need for cautious implementation and further refinement. Future research could investigate adaptive debiasing methods and optimized dataset strategies to harmonize fairness and fidelity, fostering more inclusive and dependable AI systems for societal applications.

20# References

1. Baehrens D, Schroeter T, Harmeling S, Kawanabe M, Hansen K, Müller K-R (2010) How to Explain Individual Classification Decisions. J Mach Learn Res 11:1803–1831
2. Barrio E del, Gordaliza P, Loubes J-M (2020) Review of Mathematical frameworks for Fairness in Machine Learning
3. Caton S, Haas C (2024) Fairness in Machine Learning: A Survey. ACM Comput Surv 56. https://doi.org/10.1145/3616865
4. Choi K, Grover A, Singh T, Shu R, Ermon S (2020) Fair generative modeling via weak supervision. In: Proceedings of the 37th International Conference on Machine Learning. JMLR.org
5. Chowdhury SBR, Chaturvedi S (2022) Learning Fair Representations via Rate-Distortion Maximization. Transactions of the Association for Computational Linguistics 10:1159–1174. https://doi.org/10.1162/tacl_a_00512
6. Deng J, Dong W, Socher R, Li L-J, Li K, Fei-Fei L (2009) Imagenet: A large-scale hierarchical image database. In: 2009 IEEE conference on computer vision and pattern recognition. Ieee, pp 248–255
7. Deng L (2012) The mnist database of handwritten digit images for machine learning research. IEEE Signal Processing Magazine 29:141–142
8. Dhariwal P, Nichol A (2021) Diffusion models beat GANs on image synthesis. In: Proceedings of the 35th International Conference on Neural Information Processing Systems. Curran Associates Inc., Red Hook, NY, USA
9. Elazar Y, Goldberg Y (2018) Adversarial Removal of Demographic Attributes from Text Data. In: Riloff E, Chiang D, Hockenmaier J, Tsujii J (eds) Proceedings of the 2018 Conference on Empirical Methods in Natural Language Processing. Association for Computational Linguistics, Brussels, Belgium, pp 11–21
10. Feng R, Yang Y, Lyu Y, Tan C, Sun Y, Wang C (2019) Learning Fair Representations via an Adversarial Framework. ArXiv abs/1904.13341
11. Ferrara E (2024) Fairness and Bias in Artificial Intelligence: A Brief Survey of Sources, Impacts, and Mitigation Strategies. Sci 6. https://doi.org/10.3390/sci6010003
12. Goodfellow I, Pouget-Abadie J, Mirza M, Xu B, Warde-Farley D, Ozair S, Courville A, Bengio Y (2020) Generative adversarial networks. Commun ACM 63:139–144. https://doi.org/10.1145/3422622
13. Heusel M, Ramsauer H, Unterthiner T, Nessler B, Hochreiter S (2017) GANs Trained by a Two Time-Scale Update Rule Converge to a Local Nash Equilibrium. In: Guyon I, Luxburg UV, Bengio S, Wallach H, Fergus R, Vishwanathan S, Garnett R (eds) Advances in Neural Information Processing Systems. Curran Associates, Inc.
14. Ho J, Salimans T (2021) Classifier-Free Diffusion Guidance. In: NeurIPS 2021 Workshop on Deep Generative Models and Downstream Applications
15. Hu EJ, Shen Y, Wallis P, Allen-Zhu Z, Li Y, Wang S, Wang L, Chen W (2021) LoRA: Low-Rank Adaptation of Large Language Models. arXiv preprint arXiv:210609685
16. Jain N, Olmo A, Sengupta S, Manikonda L, Kambhampati S (2022) Imperfect ImaGANation: Implications of GANs exacerbating biases on facial data augmentation and snapchat face lenses. Artificial Intelligence 304:103652. https://doi.org/10.1016/j.artint.2021.103652
17. Kamiran F, Calders T (2012) Data preprocessing techniques for classification without discrimination. Knowledge and Information Systems 33:1–33. https://doi.org/10.1007/s10115-011-0463-8

2118. Kenfack PJ, Arapovy DD, Hussain R, Kazmi SMA, Khan A (2021) On the Fairness of Generative Adversarial Networks (GANs). 2021 International Conference "Nonlinearity, Information and Robotics" (NIR) 1–7
19. Kim E, Kim S, Entezari R, Yoon S (2024) Unlocking Intrinsic Fairness in Stable Diffusion
20. Kingma DP, Welling M (2014) Auto-Encoding Variational Bayes. In: 2nd International Conference on Learning Representations, ICLR 2014, Banff, AB, Canada, April 14-16, 2014, Conference Track Proceedings
21. Krchova I, Platzer M, Tiwald P (2023) Strong statistical parity through fair synthetic data. In: NeurIPS 2023 Workshop on Synthetic Data Generation with Generative AI
22. Lee C-H, Liu Z, Wu L, Luo P (2020) MaskGAN: Towards Diverse and Interactive Facial Image Manipulation. In: IEEE Conference on Computer Vision and Pattern Recognition (CVPR)
23. Lin Y, Li D, Zhao C, Shao M (2024) Fair Data Generation via Score-based Diffusion Model. CoRR abs/2406.09495
24. Lin Y, Li D, Zhao C, Shao M, Wan G (2025) FADE: Towards Fairness-aware Augmentation for Domain Generalization via Classifier-Guided Score-based Diffusion Models
25. LUCCIONI HCS (2023) Fair Diffusion: Instructing Text-to-Image Generation Models on Fairness
26. Lundberg SM, Lee S-I (2017) A unified approach to interpreting model predictions. In: Proceedings of the 31st International Conference on Neural Information Processing Systems. Curran Associates Inc., Red Hook, NY, USA, pp 4768–4777
27. Madras D, Creager E, Pitassi T, Zemel RS (2018) Learning Adversarially Fair and Transferable Representations. CoRR abs/1802.06309
28. Maluleke VH, Thakkar N, Brooks T, Weber E, Darrell T, Efros AA, Kanazawa A, Guillory D (2022) Studying Bias in GANs Through the Lens of Race. In: Avidan S, Brostow G, Cissé M, Farinella GM, Hassner T (eds) Computer Vision – ECCV 2022. Springer Nature Switzerland, Cham, pp 344–360
29. Nicoletti L, Bass D (2023) Humans Are Biased: Generative AI Is Even Worse. https://www.bloomberg.com/graphics/2023-generative-ai-bias/
30. Oh C, Won H, So J, Kim T, Kim Y, Choi H, Song K (2022) Learning Fair Representation via Distributional Contrastive Disentanglement. In: Proceedings of the 28th ACM SIGKDD Conference on Knowledge Discovery and Data Mining. Association for Computing Machinery, New York, NY, USA, pp 1295–1305
31. Pal B, Kannan A, Kathirvel RP, O'Toole AJ, Chellappa R (2023) Gaussian Harmony: Attaining Fairness in Diffusion-based Face Generation Models
32. Parihar R, Bhat A, Basu A, Mallick S, Kundu JN, Babu RV (2024) Balancing Act: Distribution-Guided Debiasing in Diffusion Models. In: 2024 IEEE/CVF Conference on Computer Vision and Pattern Recognition (CVPR). pp 6668–6678
33. Perera MV, Patel VM (2023) Analyzing Bias in Diffusion-based Face Generation Models. In: 2023 IEEE International Joint Conference on Biometrics (IJCB). pp 1–10
34. Ramesh A, Pavlov M, Goh G, Gray S, Voss C, Radford A, Chen M, Sutskever I (2021) Zero-Shot Text-to-Image Generation. In: Meila M, Zhang T (eds) Proceedings of the 38th International Conference on Machine Learning. PMLR, pp 8821–8831
35. Ribeiro MT, Singh S, Guestrin C (2016) "Why Should I Trust You?": Explaining the Predictions of Any Classifier. In: Proceedings of the 22nd ACM SIGKDD International Conference on Knowledge Discovery and Data Mining. Association for Computing Machinery, New York, NY, USA, pp 1135–1144
36. Rombach R, Blattmann A, Lorenz D, Esser P, Ommer B (2021) High-Resolution Image Synthesis with Latent Diffusion Models. CoRR abs/2112.10752

2237. Sahili ZA, Patras I, Purver M (2025) FairCoT: Enhancing Fairness in Diffusion Models via Chain of Thought Reasoning of Multimodal Language Models
38. Salimans T, Goodfellow I, Zaremba W, Cheung V, Radford A, Chen X (2016) Improved techniques for training GANs. In: Proceedings of the 30th International Conference on Neural Information Processing Systems. Curran Associates Inc., Red Hook, NY, USA, pp 2234–2242
39. Sarhan MH, Navab N, Eslami A, Albarqouni S (2020) Fairness by Learning Orthogonal Disentangled Representations. CoRR abs/2003.05707
40. Serengil SI, Ozpinar A (2021) HyperExtended LightFace: A Facial Attribute Analysis Framework. In: 2021 International Conference on Engineering and Emerging Technologies (ICEET). IEEE, pp 1–4
41. Shen X, Du C, Pang T, Lin M, Wong Y, Kankanhalli M (2024) Finetuning Text-to-Image Diffusion Models for Fairness. In: The Twelfth International Conference on Learning Representations
42. Szegedy C, Vanhoucke V, Ioffe S, Shlens J, Wojna Z (2016) Rethinking the Inception Architecture for Computer Vision. In: 2016 IEEE Conference on Computer Vision and Pattern Recognition (CVPR). pp 2818–2826
43. Teo CTH, Abdollahzadeh M, Cheung N-M (2023) On measuring fairness in generative models. In: Proceedings of the 37th International Conference on Neural Information Processing Systems. Curran Associates Inc., Red Hook, NY, USA
44. Xie Q, Dai Z, Du Y, Hovy E, Neubig G (2017) Controllable invariance through adversarial feature learning. In: Proceedings of the 31st International Conference on Neural Information Processing Systems. Curran Associates Inc., Red Hook, NY, USA, pp 585–596
45. Xu D, Yuan S, Zhang L, Wu X (2018) FairGAN: Fairness-aware Generative Adversarial Networks. In: 2018 IEEE International Conference on Big Data (Big Data). pp 570–575
46. Zeiler MD, Fergus R (2014) Visualizing and Understanding Convolutional Networks. In: Fleet D, Pajdla T, Schiele B, Tuytelaars T (eds) Computer Vision – ECCV 2014. Springer International Publishing, Cham, pp 818–833
47. Zemel R, Wu Y, Swersky K, Pitassi T, Dwork C (2013) Learning fair representations. In: Proceedings of the 30th International Conference on International Conference on Machine Learning - Volume 28. JMLR.org, Atlanta, GA, USA, p III-325-III–333